\documentclass{article}
\usepackage{arxiv}
\usepackage[utf8]{inputenc} %
\usepackage[T1]{fontenc}    %
\usepackage[hidelinks]{hyperref}       %
\usepackage{url}            %
\usepackage{booktabs}       %
\usepackage{amsfonts}       %
\usepackage{nicefrac}       %
\usepackage{microtype}      %
\usepackage{lipsum}		%
\usepackage{graphicx}
\usepackage{threeparttable}
\usepackage{doi}
\usepackage{multirow}
\usepackage{multicol}
\usepackage{xcolor}
\usepackage{subcaption}
\usepackage[normalem]{ulem}
\usepackage{amsmath,amsfonts,bm}
\usepackage{amssymb,amsthm}
\usepackage{enumitem}
\usepackage{adjustbox}
\usepackage{blindtext}
\usepackage{cleveref}
\usepackage{float}
\usepackage{booktabs}
\usepackage{multirow}
\usepackage{array}
\usepackage{makecell} %
\usepackage[perpage]{footmisc}
\usepackage{wrapfig} 
\usepackage{tcolorbox}
\usepackage{listings}       
\usepackage{xcolor}         
\usepackage{geometry}       
\usepackage{ulem}

\usepackage[
    backend=biber,
    style=numeric,
    sorting=none,
]{biblatex}
\newcommand{\citep}[1]{\parencite{#1}}
\setlist[itemize,1]{leftmargin=\dimexpr 18pt}
\setlist[enumerate,1]{leftmargin=\dimexpr 18pt}

\makeatletter
\renewcommand{\thanks}[1]{%
  \begingroup
    \renewcommand\thefootnote{}\footnote{#1}%
    \addtocounter{footnote}{-1}%
  \endgroup
}
\makeatother

\begin{document}
\large

\title{
\raisebox{-0.1\height}{\includegraphics[width=0.04\textwidth]{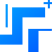}} %
Step-Audio-EditX Technical Report
}
\renewcommand{\headeright}{Preprint. Work in progress.}
\renewcommand{\undertitle}{}
\renewcommand{\shorttitle}{}
\author{
    \textbf{Chao Yan\textsuperscript{†*},} \ \textbf{Boyong Wu\textsuperscript{*},} \ \textbf{Peng Yang,} \ \textbf{Pengfei Tan,} \ \textbf{Guoqiang Hu,} \\  \textbf{Li Xie,}   \textbf{Yuxin Zhang,} \textbf{Xiangyu(Tony) Zhang,}  \textbf{Fei Tian,} \textbf{Xuerui Yang,}  \\  \textbf{Xiangyu Zhang,} \textbf{Daxin Jiang,} \textbf{Shuchang Zhou,} \textbf{Gang Yu} \\ \textbf{StepFun}
}

\thanks{* Core Contribution}
\thanks{† Corresponding Author: yanchao@stepfun.com}
\renewcommand{\thanks}[1]{\\ \large #1}

\maketitle
\thispagestyle{plain} 
\begin{abstract}
We present Step-Audio-EditX, the first open-source LLM-based reinforcement learning audio model excelling at expressive and iterative audio editing—encompassing emotion, speaking style, and paralinguistics—alongside robust zero-shot text-to-speech (TTS) capabilities. Our core innovation lies in leveraging only large-margin synthetic data in post-training, which circumvents the need for embedding-based priors or auxiliary modules. This large-margin learning approach enables both iterative control and high expressivity across voices, and represents a fundamental pivot from the conventional focus on representation-level disentanglement. Evaluation results demonstrate that Step-Audio-EditX surpasses both MiniMax-2.6-hd and Doubao-Seed-TTS-2.0 in emotion editing and other fine-grained control tasks. Our code and models are available at \url{https://github.com/stepfun-ai/Step-Audio-EditX}.

\begin{figure*}[thb]
\centering
\vspace{0.1cm} 
\includegraphics[width=0.8\textwidth]{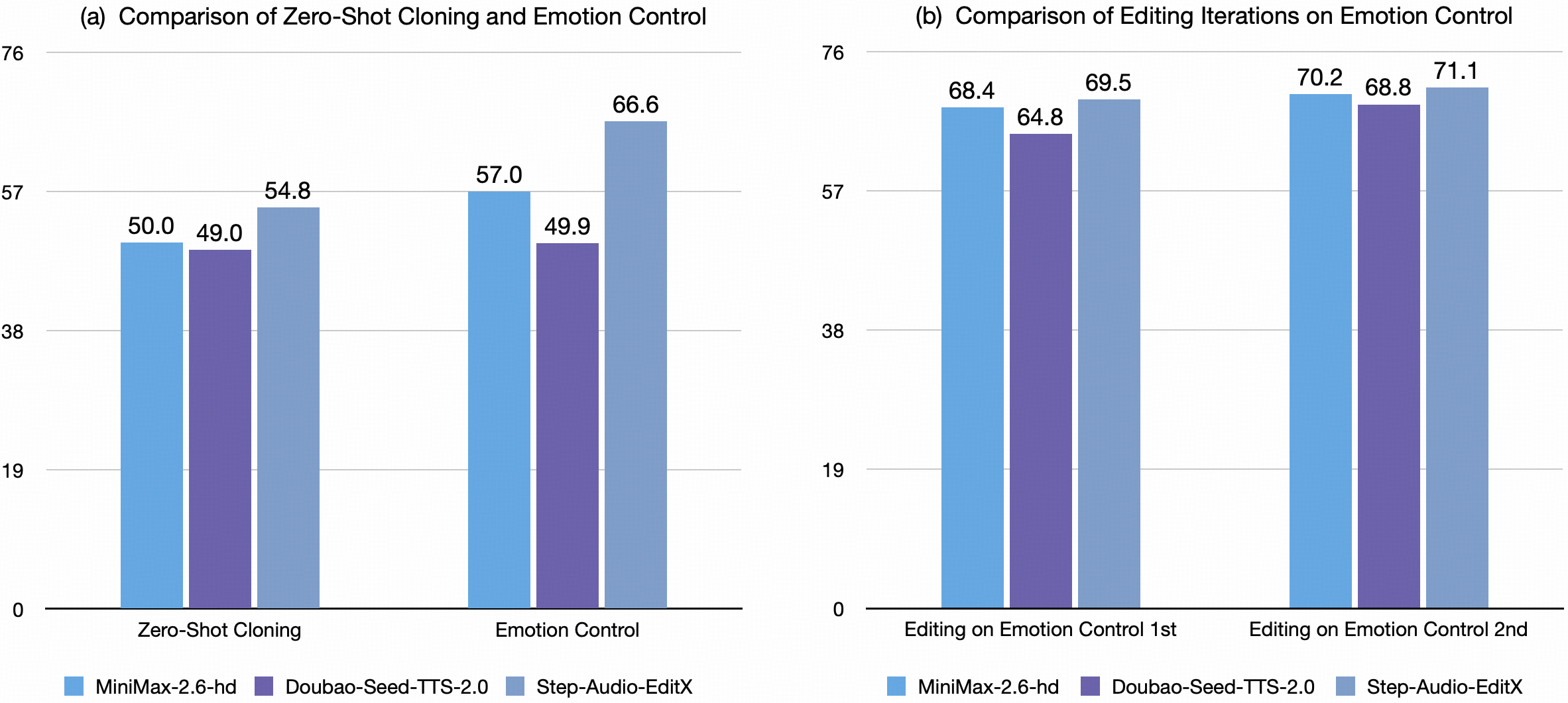}
\captionsetup{width=0.8\linewidth} 
\caption{Comparison between Step-Audio-EditX and Closed-Source models.(a) Step-Audio-EditX demonstrates superior performance over Minimax and Doubao in both zero-shot cloning and emotion control.(b) Emotion editing of Step-Audio-EditX significantly improves the emotion-controlled audio outputs of all three models after just one iteration. With further iterations, their overall performance continues to improve.
}
\label{fig:emotion}
\end{figure*}

\thispagestyle{empty}

\end{abstract}
\section{Introduction}
%

In recent years, TTS technology has advanced significantly. A notable development is zero-shot TTS models, which can generate high-quality, natural-sounding speech by mimicking the timbre, prosody, and style of a reference speech prompt. Generally, current zero-shot TTS systems fall into three main categories: those that utilize LLMs to model discrete or continuous acoustic tokens \cite{vall-e,valle2,melle,maskgct,sparktts}, those employing diffusion or flow matching models to learn direct text-to-speech mapping \cite{le2024voicebox, voiceflow, matchatts, seedtts, f5tts, f5rtts}, and hybrid coarse-to-fine systems, where LLMs first convert text tokens into coarse speech tokens, which are then refined by a diffusion or flow matching model to render fine-grained speech details\cite{cosyvoice,fireredtts,du2024cosyvoice2,fireredtts2,zhang2024speaking,du2025cosyvoice3,jia2025ditar,deng2025indextts}.

Despite considerable progress in zero-shot TTS synthesis, attributes such as emotion, style, accent, and timbre in the synthesized speech are still derived directly from the reference audio. This inherent limitation restricts independent control over these attributes. Although prepending style or emotion instructions to the input text offers a certain degree of controllability and often performs well for in-domain speakers \cite{seedtts,cosyvoice,fireredtts,du2024cosyvoice2}, this approach faces challenges in disentangling speech attributes. In particular, the cloned voice often fails to effectively follow the provided style or emotion instructions.

Many previous studies on speech disentanglement have relied on approaches such as adversarial training\cite{naturalspeech3, li2022crossspeakeremotiondisentanglingtransfer}, feature engineering\cite{choi2023dddmvcdecoupleddenoisingdiffusion,Anastassiou2024VoiceShopAU}, and innovative network architectures\cite{Jia2022ZeroShotAC} to achieve attribute decoupling. In contrast, we propose a simple yet stable data-driven method. Specifically, we design a pipeline for generating high-quality data pairs that preserve identical linguistic content while exhibiting clearly distinguishable variations in one or a few attributes, such as emotion, style, accent, and paralinguistic features. By training models on such data pairs, we achieve effective attribute disentanglement, enabling to edit the attribute of input speech. Moreover, by applying multiple iterative "editing" steps, the intensity of a target attribute can be progressively enhanced or reduced. Beyond emotion, style and paralinguistic editing, we demonstrate that this approach can be extended to other applications, including speed rate adjustment, speech denoising, and silence trimming.
In this report, we outline our contributions and findings:
\begin{itemize}
\item We present Step-Audio-EditX, the first open-source LLM-based audio model excelling at expressive and iterative audio editing, encompassing emotion, speaking style, and paralinguistics, alongside robust zero-shot TTS capabilities.
\item Our results show that emotion and speaking style can be controlled through post-training with large-margin data alone, eliminating the need for extra presentation modeling or adapter modules. 
\item We find that post-training with large-margin data enables iterative control and high expressivity across voices, which represents a fundamental pivot from conventional representation-level disentanglement methods.

\end{itemize}

\section{Architecture}
\subsection{Overview}
In our prior work, we introduced an Audio-Edit synthesis model in Step-Audio\cite{stepaudio} for nuanced emotional expressions and diverse speaking styles data generation. In this report, we retain the previous model along with the same audio tokenizer. The key modifications include an expanded range of emotions and speaking styles, the addition of zero-shot TTS and paralinguistic editing capabilities, as well as a reduction in model parameters from 130B to 3B. Leveraging large-margin synthetic data, our 3B model demonstrates superior and more stable performance compared to the previous version.
\\Our system comprises three primary components: (1) a dual-codebook audio tokenizer, which converts reference or input audio into discrete tokens; (2) an audio LLM that generates dual-codebook token sequences; and (3) an audio decoder, which converts the dual-codebook token sequences predicted by the audio LLM back into audio waveforms using a flow matching approach.This integrated architecture enables the Step-Audio-EditX to perform zero-shot TTS and diverse editing tasks within a unified framework. Thus, it can directly capitalize on the rich ecosystem of post-training techniques developed for text LLMs.

\begin{figure*}[thb]
\centering
\vspace{0.1cm} 
\includegraphics[width=1\textwidth]{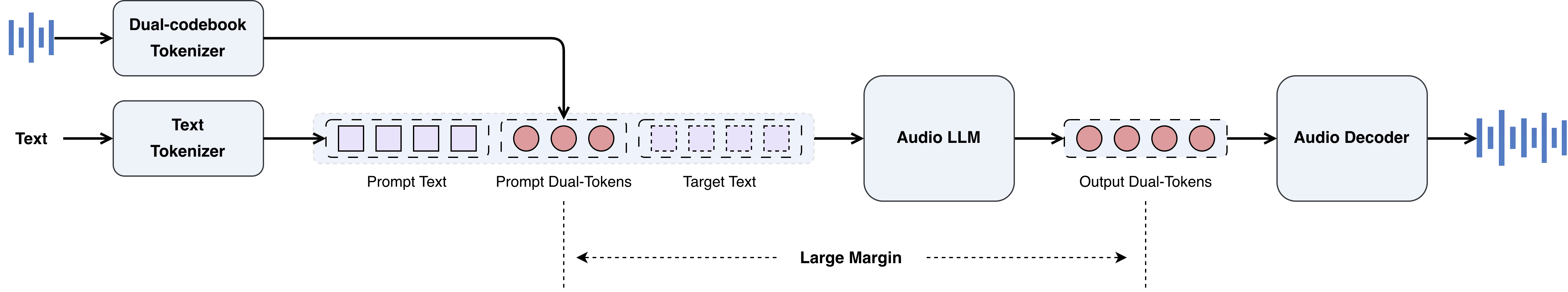}
\vspace{-0.2cm}
\caption{An overview of the architechture of Step-Audio-EditX}
\label{fig:overall}
\end{figure*}

\subsection{Audio Tokenizer}
We investigate the effect of LLMs post-training with large-margin data by retaining the dual-codebook tokenization framework from our previous Step-Audio model, which employs parallel linguistic (16.7 Hz, 1024-codebook) and semantic (25 Hz, 4096-codebook) tokenizers in a 2:3 interleaving ratio. Based on a series of downstream audio tokenizer reconstruction experiments, we observed that the dual-codebook tokenizer retains a considerable amount of emotional, prosodic, and other non-linguistic information, indicating suboptimal disentanglement. This shortcoming makes it particularly suitable for validating the effectiveness of our LLM post-training strategy and the proposed large-margin data-driven methodology.

\subsection{Audio LLM}
The audio LLM uses the same architecture as our prior Audio-Edit model, differing only in its smaller parameter size of 3B. To capitalize on the powerful language capabilities of pre-trained text LLMs, the 3B model is initialized with a text-based LLM, followed by training on a blended dataset with a 1:1 ratio of text data to audio dual-codebook tokens. The audio LLM processes text tokens along with their corresponding dual-codebook audio tokens in a chat format, subsequently generating dual-codebook tokens as the sole output.

\subsection{Audio Decoder}
The audio decoder consists of a Flow Matching module and a BigVGANv2\cite{lee2023bigvganuniversalneuralvocoder} vocoder. The flow matching module generates Mel spectrograms, given output audio tokens, reference audio, and speaker embedding as conditions, while the BigVGANv2 vocoder further converts the Mel spectrograms into waveforms. For the flow matching module, we adopt the diffusion transformer (DiT) as the backbone,  and train the model on 200,000 hours of high-quality speech. This enhancement significantly improves its Mel spectrogram reconstruction capability, leading to substantial gains in both pronunciation accuracy and timbre similarity.

\section{Data}
Consistent with prior work on StepAudio's pre-training dataset and methodology, this report focus on the post-training dataset and the corresponding methods.

\subsection{SFT Data}
We employ SFT to enable the Step-Audio-EditX model for zero-shot TTS and diverse audio editing tasks. The SFT data can be categorized into several parts: zero-shot TTS, emotion editing, speaking style editing and paralinguistic editing. Notably, the large-margin dataset targets editing tasks, particularly on the aspects of emotion and speaking style.

\subsubsection{Zero-shot Text to Speech}
We employ a high-quality, professionally annotated in-house dataset, primarily in Chinese and English, for zero-shot TTS. Furthermore, a minimal amount of Cantonese and Sichuanese data is employed to elicit dialect capabilities. To ensure diverse and highly expressive styles and emotions in the synthesized speech, as well as robust zero-shot performance, The dataset captures vocal variations within individual speakers as well as across a broad speaker population, comprising approximately 60,000 unique individuals.

\subsubsection{Emotion and Speaking Style Editing}
Emotion and speaking style present significant challenges for expressive text-to-speech systems, due to the inherent difficulties in both defining their categorical characteristics and collecting high-quality data. We propose a straightforward and efficient large-margin synthetic data approach, which performs zero-shot voice cloning across different emotions and speaking styles for the same speaker, while ensuring a sufficiently large margin between contrastive sample pairs. Only one prompt audio segment per emotion or speaking style is required, eliminating the need for costly data collection. Moreover,  this method ingeniously converts complex emotion and style descriptions into a comparative pair-based data construction format. In the following, we introduce the proposed approach:

\textbf{Voice Actor Recording.} Voice actors recorded expressive emotions and speaking styles. For each actor, a single audio clip of approximately 10 seconds was captured for every emotion and style combination.

\textbf{Zero-shot Cloning.} A triplet $\langle text\textsubscript{prompt},audio\textsubscript{neutral}, audio\textsubscript{emotion,style}\rangle$  is constructed for each emotion and speaking style by selecting corresponding emotional and neutral audio clips from the same speaker as the prompt audio and processing them with the StepTTS voice cloning interface, using a text instruction that describes the target attribute.

\textbf{Margin Scoring.} To evaluate the triplet generated, we developed a scoring model using a small, human-annotated dataset. The model evaluates audio pairs on a 1-10 scale, with higher margin scores corresponding to more desirable outcomes.

\textbf{Margin Selection.} Samples were selected based on a margin score threshold. This threshold was adjusted for different emotions and styles, with a score of 6 serving as the universal lower bound.

Notably, the audio clips in each triplet are generated using the same emotional or stylistic text prompt, which encourages the model to focus solely on the variations in emotion and style itself during the SFT training process.

\subsubsection{Paralinguistic Editing}
Paralinguistic cues, such as breathing, laughter, and filled pauses (e.g., "uhm"), are crucial for enhancing the naturalness and expressiveness of synthesized speech. We achieved paralinguistic editing capability by using a "semi-synthetic" strategy, which leverages the NVSpeech dataset\cite{liao2025nvspeech}, a highly expressive speech corpus whose rich annotations for numerous paralinguistic types enabled the construction of comparative quadruplets for model training. The quadruplet $\langle text\textsubscript{without\_tags}, audio\textsubscript{without\_tags}, text\textsubscript{nv\_source}, audio\textsubscript{nv\_source}\rangle$ construction differs from the triplet by using the NVSpeech original audio and transcript as the target output and the StepTTS voice cloning generated audio as the input, which is synthesized using the original transcript after paralinguistic tag removal.

As paralinguistic editing is an editing task performed in the time domain and exhibits substantial intrinsic margin differences, a margin scoring model is not required for data selection. A small set of quadruplet data is sufficient to effectively elicit the model's paralinguistic editing capabilities.

\subsection{Reinforcement Learning Data}
To align our model with human preferences, we construct two distinct types of preference datasets using different approaches: one based on human annotation, and the other employing the LLM-as-a-Judge method.

\textbf{Human Annotation.} We first collected real-world prompt audio and corresponding text prompts from users, and generated 20 candidate responses using the SFT model. We then constructed chosen/rejected pairs by having human annotators rate each of the 20 responses on a 5-point scale based on the criteria of correctness, prosody, and naturalness. Only pairs with a score margin greater than 3 were selected.

\textbf{LLM-as-a-Judge.} Model responses were scored on a 1-10 scale for emotion and speaking style editing by a comprehension model. Preference pairs were then generated from these scores, retaining only pairs with a score margin greater than 8 points in the final dataset. 

These selected large-margin pairs will be used to train the reward model and PPO.

\section{Training}
Our post-training process aligns the model’s outputs with zero-shot TTS, a variety of editing tasks, and human preferences. This alignment is accomplished through a two-stage approach: SFT followed by proximal policy optimization.

\subsection{ Supervised Fine-tuning}
The SFT stage enhances the model's zero-shot text-to-speech synthesis and editing capabilities through the use of distinct system prompts in a chat format. In the zero-shot TTS task, the prompt waveform is encoded into dual-codebook tokens, which are subsequently detokenized into string format and incorporated into the speaker information within the system prompt. The text to be synthesized serves as the user prompt in a chat-based format, and the generated dual-code tokens are returned as the system's response. For the editing task, all operations are defined under a unified system prompt. The user prompt includes both the original audio and a descriptive command for the editing operation, and the system response delivers the resulting edited audio tokens. The model is finetuned for one epoch with learning rate from $1 \times 10^{-5}$ to $1 \times 10^{-6}$.

\subsection{Reinforcement Learning} 
Reinforcement learning has further amplified the model's stability in zero-shot TTS, as well as its capability and expressiveness in following editing instructions. These enhancements are particularly noticeable when there is a substantial divergence between the emotional and stylistic characteristics of the source prompt waveform and the target editing output, such as generating sad speech from a happy prompt or converting loud speech into a whisper. This reinforcement learning approach offers a novel perspective to address these challenges by shifting the focus from achieving ideal speech representation disentanglement to improving both the construction of large-margin pairs and the efficacy of the reward model evaluation.

\textbf{Reward Model Training.} The reward model is initialized from a 3B SFT model and is trained using a combination of human-annotated and LLM-as-a-judge-generated large-margin data, optimized with the Bradley-Terry loss. The model is a token-level reward model trained directly on large-margin dual-codebook token pairs. This approach obviates the need to convert tokens back into waveform using an audio decoder during reward computation. The model is finetuned for one epoch and the learning rate is adjusted using a cosine decay strategy, initialized at $2 \times 10^{-5}$ with a lower bound set at $1 \times 10^{-5}$.

\textbf{PPO Training.} Following the acquisition of the reward model, we employ the PPO algorithm for further training, using the same prompt seeds as in the reward model training, except for the selection of only the most challenging prompts for the SFT model. In the PPO training stage, the critic model is warmed up for 80 steps ahead of the actor. The optimizer uses an initial learning rate of $1 \times 10^{-6}$, which follows a cosine decay schedule with a lower bound of $2 \times 10^{-7}$. A PPO clip threshold of $\epsilon = 0.2$ and a KL divergence penalty with a coefficient of $\beta = 0.05$ are applied.

\section{Evaluation}
The accurate and comprehensive evaluation of a model’s performance in synthesizing emotional, stylistic, and paralinguistic speech represents a substantial challenge. To address this, we first introduce the construction of a comprehensive and reproducible benchmark in Section 5.1. We then employ this benchmark in Section 5.2 to demonstrate the advantages of our Step-Audio-EditX model.

\subsection{Evaluation Benchmark}

We introduce Step-Audio-Edit-Benchmark, a benchmark that leverages LLM-as-a-judge model to evaluate model performance on emotion, speaking style, and paralinguistics. All evaluation audio is generated via zero-shot voice cloning and subsequently scored using the Gemini-2.5-Pro\footnote{https://blog.google/products/gemini/gemini-2-5-pro-latest-preview/} model.

\textbf{Speaker Selection.} The speaker set for zero-shot cloning consisted of eight speakers (2 male and 2 female per language, for both Chinese and English). The Chinese speakers were sourced from the Wenet-Speech4TTS\cite{ma2024wenetspeech4tts12800hourmandarintts} corpus, whereas the English speakers were sourced from the open-source GLOBE-V2\cite{wang2024globehighqualityenglishcorpus} and Libri-Light\cite{Kahn_2020} datasets, respectively.

\textbf{Emotion.} The emotional test set covers five categories: happiness, anger, sadness, fear, and surprise. Each category includes 50 Chinese and 50 English prompts, with the textual content of each prompt designed to be consistent with its corresponding target emotion.

\textbf{Speaking Style.} The test set includes seven speaking styles: childlike, elderly, exaggerated, recitative, passionate, coquettish, and whisper. Each style contains 50 Chinese and 50 English prompts, with content matched to its target style.

\textbf{Paralinguistic.} The paralinguistic test set includes ten paralinguistic labels per speaker: breathing, laughter, surprise-oh, confirmation-en, uhm, surprise-ah, surprise-wa, sigh, question-ei, and dissatisfaction-hnn. Each label contains 50 relevant LLM-generated samples in Chinese and 50 in English.

\textbf{Emotion and Speaking Style Evaluation.} To evaluate emotion and speaking style, predefined category sets (5 emotions and 7 styles) are provided to the Gemini-2.5-Pro model in the prompts, instructing it to classify the audio. The final accuracy for each category is calculated as the average across all speakers.

\textbf{Paralinguistic Style Evaluation.} To evaluate the performance of paralinguistic editing, a specialized evaluation prompt has been designed for the Gemini-2.5-Pro model, employing a rigorous 1–3 scoring scale (3 = perfect, 2 = flawed, 1 = failed). The prompt directs the model to actively examine specific assessment points in the audio—such as whether annotations like [laughter] or [sigh] have been accurately inserted. Particular emphasis is placed on the most common failure mode, “omission,” where the audio may remain fluent but lacks required paralinguistic elements specified in the instructions. Finally, model performance in the paralinguistic editing task is assessed by calculating the overall average score generated by Gemini-2.5-Pro model.

\subsection{Evaluation Results}
This section details our model's performance on the Step-Audio-Edit-Benchmark, benchmark and demonstrates its superior editing accuracy and scalability when used to edit audio generated by various closed-source TTS systems.

\subsubsection{Emotion and Speaking Style Editing Results}

This evaluation employs an iterative approach to audio editing for emotion and speaking style. The process begins with a zero-shot clone as the initial audio $iteration_0$, which then undergoes $N$ rounds of iterative editing. The output of the N-th round is denoted as $iteration_N$. In this specific setup, N is configured as 3. For most use cases, two editing iterations are sufficient to meet the desired criteria. 

\textbf{Iterative Editing Results.} As shown in Table~\ref{tab:emotion_style_accuracy}, there was a significant boost in both emotion and speaking style accuracy after the initial edit of the Iter\textsubscript{0} audio. Furthermore, with successive iterations of editing, the accuracy for both emotion and speaking style was further enhanced.

\begin{table}[ht]
\centering
\caption{Performance of Step-Audio-EditX on Emotion and Speaking Style Editing.}
\label{tab:emotion_style_accuracy}
\begin{tabular}{lcccc|cccc}
\toprule
& \multicolumn{4}{c}{\textbf{Emotion}\  $\uparrow$} & \multicolumn{4}{c}{\textbf{Speaking Style}\ $\uparrow$} \\
\cmidrule(lr){2-5} \cmidrule(lr){6-9}
\textbf{Language} & \textbf{Iter\textsubscript{0}} & \textbf{Iter\textsubscript{1}} & \textbf{Iter\textsubscript{2}} & \textbf{Iter\textsubscript{3}} & \textbf{Iter\textsubscript{0}} & \textbf{Iter\textsubscript{1}} & \textbf{Iter\textsubscript{2}} & \textbf{Iter\textsubscript{3}} \\
\midrule
Chinese & 58.7 & 73.6 & 75.1 & \textbf{77.8} & 40.4 & 62.1 & 65.3 & \textbf{68.0} \\ 
English & 51.2 & 60.0 & 63.1 & \textbf{64.2} & 48.8 & 63.4 & 62.3 & \textbf{64.4} \\
\underline{Average} & 55.0 & 66.8 & 69.1 & \textbf{71.0} & 44.6 & 62.8 & 63.8 & \textbf{66.2} \\
\midrule
Chinese (Prompt-Fixed) & 57.5 & 73.1 & \textbf{76.3} & 75.8 & 41.1 & 62.0 & \textbf{65.1} & 63.7 \\
English (Prompt-Fixed) & 49.7 & 60.4 & 61.1 & \textbf{62.8} & 50.0 & 63.4 & 63.2 & \textbf{63.9} \\
\underline{Average} & 53.6 & 66.8 & 68.7 & \textbf{69.3} & 45.6 & 62.7 & 64.2 & \textbf{63.8} \\
\bottomrule
\end{tabular}
\end{table}
\textbf{Prompt Audio Ablation.} Since the performance improvement in later iterations (starting from Iter\textsubscript{2}) were attributed to both the dual-code and the prompt audio. To isolate the effect of the prompt audio, we conducted an ablation study in which the prompt audio was held constant across all iterations. As presented in the Prompt-Fixed section of Table~\ref{tab:emotion_style_accuracy}, the accuracy for both emotion and speaking style continues to improve with an increasing number of editing iterations. This clearly demonstrates the effectiveness of our large-margin method. 

\textbf{\begin{table}[ht]
\centering
\caption{Generalization of Emotion and Speaking Style Editing on Closed-Source Models.}
\label{tab:api_accuracy}
\begin{tabular}{llcccc|cccc}
\toprule
& & \multicolumn{4}{c}{\textbf{Emotion} \ $\uparrow$} & \multicolumn{4}{c}{\textbf{Speaking Style} \ $\uparrow$} \\
\cmidrule(lr){3-6} \cmidrule(lr){7-10}
\textbf{Language} & \textbf{Model} & \textbf{Iter\textsubscript{0}} & \textbf{Iter\textsubscript{1}} & \textbf{Iter\textsubscript{2}} & \textbf{Iter\textsubscript{3}} & \textbf{Iter\textsubscript{0}} & \textbf{Iter\textsubscript{1}} & \textbf{Iter\textsubscript{2}} & \textbf{Iter\textsubscript{3}} \\
\midrule
\multirow{4}{*}{Chinese} & MiniMax-2.6-hd & 71.6 & 78.6 & 81.2 & \textbf{83.4} & 36.7 & 58.8 & 63.1 & \textbf{67.3} \\
      & Doubao-Seed-TTS-2.0 & 67.4 & 77.8 & 80.6 & \textbf{82.8} & 38.2 & 60.2 & \textbf{65.0} & 64.9 \\
      & GPT-4o-mini-TTS & 62.6 & 76.0 & 77.0 & \textbf{81.8} & 45.9 & 64.0 & 65.7 & \textbf{69.7} \\
      & ElevenLabs-v2 & 60.4 & 74.6 & 77.4 & \textbf{79.2} & 43.8 & 63.3 & 69.7 & \textbf{70.8} \\
\midrule
\multirow{4}{*}{English} & MiniMax-2.6-hd & 55.0 & 64.0 & 64.2 & \textbf{66.4} & 51.9 & 60.3 & 62.3 & \textbf{64.3} \\
      & Doubao-Seed-TTS-2.0 & 53.8 & 65.8 & 65.8 & \textbf{66.2} & 47.0 & 62.0 & \textbf{62.7} & 62.3 \\
      & GPT-4o-mini-TTS & 56.8 & 61.4 & 64.8 & \textbf{65.2} & 52.3 & 62.3 & 62.4 & \textbf{63.4} \\
      & ElevenLabs-v2 & 51.0 & 61.2 & 64.0 & \textbf{65.2} & 51.0 & 62.1 & 62.6 & \textbf{64.0} \\
\midrule
\multirow{4}{*}{\underline{Average}} & MiniMax-2.6-hd & 63.3 & 71.3 & 72.7 & \textbf{74.9} & 44.2 & 59.6 & 62.7 & \textbf{65.8} \\
      & Doubao-Seed-TTS-2.0 & 60.6 & 71.8 & 73.2 & \textbf{74.5} & 42.6 & 61.1 & \textbf{63.9} & 63.6 \\
      & GPT-4o-mini-TTS & 59.7 & 68.7 & 70.9 & \textbf{73.5} & 49.1 & 63.2 & 64.1 & \textbf{66.6} \\
      & ElevenLabs-v2 & 55.7 & 67.9 & 70.7 & \textbf{72.2} & 47.4 & 62.7 & 66.1 & \textbf{67.4} \\
\bottomrule
\end{tabular}
\end{table}}

\textbf{Generalization on Closed-Source Models.} The emotion and speaking style generalization of the Step-Audio-EditX model was evaluated on several leading closed-source TTS systems, including GPT-4o-mini-TTS\footnote{https://platform.openai.com/docs/guides/text-to-speech}, Eleven\_Multilingual\_v2\footnote{https://elevenlabs.io/docs/api-reference/text-to-speech/convert}, Doubao-Seed-TTS-2.0\footnote{https://www.volcengine.com/docs/6561/1871062}, and MiniMax-speech-2.6-hd\footnote{https://platform.minimaxi.com/docs/api-reference/speech-t2a-http}. For each TTS system, one male and one female built-in voice were selected for direct speech synthesis of the source text. Subsequently, three iterations of editing were applied to the resultant audio outputs. As presented in Table~\ref{tab:api_accuracy}, the built-in voices of these closed-source systems possess considerable in-context capabilities, allowing them to partially convey the emotions in the text. After a single editing round with Step-Audio-EditX, the emotion and style accuracy across all voice models exhibited significant improvement. Further enhancement was observed over the next two iterations, robustly demonstrating our model's strong generalization.

\textbf{Emotion Control on Closed-Source Models.} Due to the limited availability of closed-source systems with emotion and speaking style control, this section presents a comparative evaluation of Doubao-Seed-TTS-2.0 and MiniMax-speech-2.6-hd, selected for their capability in both zero-shot cloning and emotion control. To meet the minimum audio length constraints of the closed-source models and to ensure a fair evaluation, the prompt audios for all the speakers in the Step-Audio-Edit-Benchmark were extended in duration. These extended audios were employed for zero-shot cloning followed by two emotion editing iterations. Additionally, the cloned voices were used to generate emotional speech via each closed-source model's native emotion control.The outputs from this native emotion control subsequently underwent one round of editing with our model. It can be observed from Table~\ref{tab:api_accuracy_2} that:
\begin{itemize}
\item Our Step-Audio-EditX demonstrates better emotional accuracy in its zero-shot cloning capability compared to the other two models. 
\item The emotional accuracy of all audio samples was significantly improved after just one editing iteration. 
\item One emotional editing iteration applied to the zero-shot cloned audio outperformed the results generated by the native emotion control functions of the closed-source models.
\end{itemize}
\begin{table}[ht]
\centering
\caption{Performance Comparison Between Step-Audio-EditX and Closed-Source Models on Emotion Editing.}
\label{tab:api_accuracy_2}
\begin{tabular}{llcccc}
\toprule
& & \multicolumn{4}{c}{\textbf{Emotion}\  $\uparrow$} \\
\cmidrule(lr){3-6} 
\textbf{Language} & \textbf{Model} & \textbf{Iter\textsubscript{0}} & \textbf{Iter\textsubscript{1}} & \textbf{Iter\textsubscript{2}} & \textbf{Iter\textsubscript{3}}  \\
\midrule
\multirow{6}{*}{Chinese} & Step-Audio-EditX & 58.6 & 72.1 & 75.7 & \textbf{77.8}  \\
      & MiniMax-2.6-hd (Clone) & 49.4 & 72.1 & 75.6 & \textbf{78.1} \\
      & MiniMax-2.6-hd (Emotion Control) & - & 59.9 & 75.2 & \textbf{78.2} \\
      & Doubao-Seed-TTS-2.0 (Clone) & 50.8 & 70.9 & 75.6 & \textbf{76.4} \\
      & Doubao-Seed-TTS-2.0 (Emotion Control) & - & 51.8 & 68.9 & \textbf{75.9} \\
\midrule
\multirow{6}{*}{English} & Step-Audio-EditX & 51.0 & 61.0 & 63.2 & \textbf{64.3}  \\
      & MiniMax-2.6-hd (Clone) & 50.6 & 60.2 & 62.9 & \textbf{63.6}\\
      & MiniMax-2.6-hd (Emotion Control) & - & 54.0 & 61.5 & \textbf{62.1}  \\
      & Doubao-Seed-TTS-2.0 (Clone) & 47.1 & 57.7 & 61.9 & \textbf{64.5}  \\
      & Doubao-Seed-TTS-2.0 (Emotion Control) & - & 47.9 & 60.7 & \textbf{61.7}  \\
\midrule
\multirow{6}{*}{\underline{Average}} & Step-Audio-EditX & 54.8 & 66.6 & 69.5 & \textbf{71.1} \\
      & MiniMax-2.6-hd (Clone) & 50.0 & 66.2 & 69.3 & \textbf{70.9} \\
      & MiniMax-2.6-hd (Emotion Control) & - & 57.0 & 68.4 & \textbf{70.2} \\
      & Doubao-Seed-TTS-2.0 (Clone) & 49.0 & 64.3 & 68.8 & \textbf{70.5}  \\
      & Doubao-Seed-TTS-2.0 (Emotion Control) & - & 49.9 & 64.8 & \textbf{68.8}  \\
\bottomrule
\end{tabular}
\end{table}

\subsubsection{Paralinguistic Results}
Paralinguistic editing can be considered a time-domain operation. We evaluated the effect of a single editing iteration using Step-Audio-EditX and assessed its generalization across other closed-source models.

\textbf{Paralinguistic Editing Results.} As shown in Table~\ref{tab:para-tab1}, a significant performance gain is obtained by adding paralinguistic tags in a single editing iteration.

\begin{table}[ht]
\centering
\caption[Performance of Step-Audio-EditX on Paralinguistic Editing]{%
Performance of Step-Audio-EditX on Paralinguistic Editing.\\
(Evaluated by LLM-Judge on a 1-3 scale)%
}
\label{tab:para-tab1}
\begin{tabular}{lccc}
\toprule
& \multicolumn{2}{c}{\textbf{Paralinguistic}\  $\uparrow$} \\
\cmidrule(lr){2-3} 
\textbf{Language} & \textbf{Iter\textsubscript{0}} & \textbf{Iter\textsubscript{1}} \\
\midrule
Chinese & 1.80 &  2.89 \\
English & 2.02 &  2.89 \\
\midrule
\underline{Average} & 1.91 &  2.89 \\
\bottomrule
\end{tabular}
\end{table}

\textbf{Generalization on Closed-Source Models.} The generalization evaluation was conducted identically to the prior one. For each closed-source model, we employed one female and one male built-in voice to synthesize speech from texts with paralinguistic labels removed. The resultant audio then underwent a single editing iteration. Additionally, for comparison, extra audio samples were synthesized by substituting paralinguistic tags with onomatopoeic words (e.g., "[Laughter]" → "haha"). After one iteration of paralinguistic editing with Step-Audio-EditX, the performance of paralinguistic reproduction is comparable to that achieved by the built-in voices of closed-source models when synthesizing native paralinguistic content directly.

\begin{table}[ht]
\centering
\caption[Generalization of Paralinguistic Editing on Close-Source Models]{%
Generalization of Paralinguistic Editing on Close-Source Models.\\
(Evaluated by LLM-Judge on a 1-3 scale)%
}
\label{tab:api_generalization_combined}
\begin{tabular}{llccc}
\toprule
& & \multicolumn{3}{c}{\textbf{Paralinguistic}\  $\uparrow$} \\
\cmidrule(lr){3-5} 
\textbf{Language} & \textbf{Model} & \textbf{Iter\textsubscript{0}} & \textbf{Substitution} & \textbf{Iter\textsubscript{1}} \\
\midrule
\multirow{4}{*}{Chinese}
& MiniMax-speech-2.6-hd & 1.73 & 2.80 & 2.90 \\
& Doubao-Seed-TTS-2.0 & 1.67 & 2.81 & 2.90 \\
& GPT-4o-mini-TTS & 1.71 & 2.88 & 2.93 \\
& ElevenLabs-v2 & 1.70 & 2.71 & 2.92 \\
\midrule
\multirow{4}{*}{English}
& MiniMax-speech-2.6-hd & 1.72 & 2.87 & 2.88 \\
& Doubao-Seed-TTS-2.0 & 1.72 & 2.75 & 2.92 \\
& GPT-4o-mini-TTS & 1.90 & 2.90 & 2.88 \\
& ElevenLabs-v2 & 1.93 & 2.87 & 2.88 \\
\midrule
\multirow{4}{*}{\underline{Average}}
& MiniMax-speech-2.6-hd & 1.73 & 2.84 & 2.89 \\
& Doubao-Seed-TTS-2.0 & 1.70 & 2.78 & 2.91 \\
& GPT-4o-mini-TTS & 1.81 & 2.89 & 2.90 \\
& ElevenLabs-v2 & 1.82 & 2.79 & 2.90 \\
\bottomrule
\end{tabular}
\end{table}

The evaluation results across emotion, speaking style, and paralinguistic editing tasks confirm that our simple yet powerful approach—large-margin learning with reinforcement learning enhancement—delivers high accuracy and strong generalization. This methodology demonstrates considerable promise for both advancing research and enabling practical applications.

\section{Extensions}
This large-margin learning method can be straightforwardly extended to various downstream applications. By enforcing a sufficiently large margin between paired data samples, the model can rapidly acquire target editing capabilities through SFT. Reinforcement learning can then be seamlessly integrated to further enhance performance on challenging cases. This section details two practical extensions: (1) speed editing for speech rate control, and (2) denoising and silence trimming.

\subsection{Speed Editing}
Speed editing addresses the need for adjustable speech rates across different speakers and scenarios. This is achieved by constructing $\langle text, audio\textsubscript{source}, audio\textsubscript{faster,slower}\rangle$ triplet, where the speed-modified versions for a given speaker are generated through controlled speed perturbation using the SoX-toolkit\cite{soxtoolkit}. Since speech rate variations directly lead to substantial disparities in token sequence lengths, even SFT alone is sufficient for effective speed editing.

\subsection{Denoising and  Silence Trimming}
Background noise and silence segments in prompt audio can substantially influence the performance of zero-shot voice cloning. The model tends to interpret these acoustic features as part of the speaker's characteristics, subsequently reproducing them in synthesized audio. While such imitation is desirable in some use cases, it is undesirable in others. To address this, we integrated denoising and silence trimming using a generative approach, which enables targeted editing of both the prompt and the synthesized audio. 

\textbf{Denoising.} The triplet $\langle text, audio\textsubscript{augment}, audio\textsubscript{source}\rangle$ is constructed for denoising, with audio\textsubscript{source} serving as the ground-truth reference and audio\textsubscript{augment} being generated through additive noise and reverberation simulation.

\textbf{Silence Trimming.} The triplet is defined as $\langle text, audio\textsubscript{source}, audio\textsubscript{trimming}\rangle$, where audio\textsubscript{source} corresponds to the source audio containing silent segments, and audio\textsubscript{trimming} refers to the processed version generated by extracting and concatenating speech segments according to the timestamps produced by Silero-VAD\cite{SileroVAD}.

\section{Conclusion}
In this work, we present Step-Audio‑EditX, an LLM-based audio model trained on large-margin data and enhanced through reinforcement learning. The model enables zero-shot TTS, iterative editing of emotion and speaking style, and paralinguistic editing. We have identified that the capabilities of LLMs and the use of large-margin data, which have often been overlooked in previous studies, allow the model to overcome the limitations of audio representations. Furthermore, the proposed framework can be easily extended to a variety of tasks, including dialect editing, accent editing, vocal editing, and imitation. Finally, it should be noted that our audio editing process is not strictly conventional "editing" in the traditional sense. Instead, it functions as a form of conditional regeneration or transfer. For tasks that require partial modifications while preserving the rest of the content, our approach provides a straightforward yet effective mask-based editing method by reconstructing paired data to ensure only specific portions of the edited tokens differ from the original sequence.

\setlength{\bibitemsep}{0.5\baselineskip} 

\printbibliography[title={References}]

\newpage
\appendix


\end{document}